\documentclass{article}

\usepackage[preprint]{neurips_2026}


\usepackage[utf8]{inputenc} % allow utf-8 input
\usepackage[T1]{fontenc}    % use 8-bit T1 fonts
\usepackage{hyperref}       % hyperlinks
\usepackage{url}            % simple URL typesetting
\usepackage{booktabs}       % professional-quality tables
\usepackage{amsfonts}       % blackboard math symbols
\usepackage{nicefrac}       % compact symbols for 1/2, etc.
\usepackage{microtype}      % microtypography
\usepackage{xcolor}         % colors

\usepackage{amsmath}
\usepackage[capitalise]{cleveref}
\usepackage{amssymb}
\usepackage{algorithm}
\usepackage{algpseudocode}
\usepackage{graphicx}

\newtheorem{proposition}{Proposition}

\newtheorem{remark}{Remark}

\title{Source Distribution Estimation by Posterior Averaging}

\author{%
  Trung-Dung Hoang \\
  Department of Digital Medicine, University of Bern, Switzerland \\
  Department of Diabetes, Endocrinology, Nutritional Medicine and Metabolism UDEM, \\ Inselspital, Bern University Hospital, University of Bern, Switzerland \\
  Graduate School for Cellular and Biomedical Sciences (GCB), University of Bern, Switzerland \\
  Diabetes Center Berne, Switzerland \\
  \texttt{trung.hoang@unibe.ch} \\
  \And
  Lisa M. Koch \\
  Department of Digital Medicine, University of Bern, Switzerland \\
  Department of Diabetes, Endocrinology, Nutritional Medicine and Metabolism UDEM, \\ Inselspital, Bern University Hospital, University of Bern, Switzerland \\
  Diabetes Center Berne, Switzerland \\
  \texttt{lisa.koch@unibe.ch}
}

\begin{document}

\maketitle

\begin{abstract}
Simulation-based science often requires a distribution over simulator parameters whose push-forward reproduces a set of real observations: this is the source distribution estimation (SDE) problem. Existing methods fit the source against a likelihood surrogate trained once from a fixed proposal prior. Their objective is therefore stated only in terms of the surrogate instead of the true simulator, which may fail for inaccurate areas in parameter space where the surrogate was never trained. We instead solve SDE by expectation maximization: an E-step trains an amortized posterior on fresh simulations from the current source estimate, and an M-step refits the source to the average of that posterior over the observed data. 
We give two parameterizations, (1) separate source and posterior flows and (2) a single shared conditional flow. We evaluate our method on three benchmark tasks under both broad and misspecified initial priors. Both improve on existing fixed surrogate approaches and on iterated variants of each, most clearly on Lotka--Volterra, where no baseline falls below 0.96 data-space C2ST while our methods reach 0.64-0.68 in three of four initial-prior settings. 
\end{abstract}

\section{Introduction}
\label{sec:intro}
Researchers often encode their knowledge of a system as a \textit{simulator} in many domains such as astrophysics, epidemiology, or neuroscience: given parameters, one can generate synthetic observations and study how the system works internally \citep{sbi_frontier}. However, what is typically available is not the distribution of parameters but a distribution of real observations. Therefore, to run the simulator consistently with reality, one must infer a distribution over parameters that is compatible with the data. This identification problem is known as \emph{Source Distribution Estimation (SDE)} \citep{vandegar2021neural,vetter2024sourcerer}. Formally, the simulator is represented as $p(x \mid \theta)$, where, given parameter $\theta$, we can sample $x \sim p(x \mid \theta)$ via simulation, but the density $p(x \mid \theta)$ itself is intractable. Given a dataset $\mathcal{D} = \{x_1, \dots, x_n\}$ of real observations with empirical distribution $p_o(x)$, we need to find the source distribution $q(\theta)$ such that $q(x) = \int p(x \mid \theta)\, q(\theta) \, d\theta$ matches $p_o(x)$.
Throughout this paper we assume the simulator $p(x \mid \theta)$ is a \textit{black box}: we can draw forward samples $x \sim p(x \mid \theta)$ for any $\theta$, but we have no access to the simulator's implementation or gradients $\nabla_\theta p(x \mid \theta)$. This is the common setting for licensed, proprietary scientific simulators ~\citep{YAO2026140531}, and for simulators with discrete or combinatorial internal stochasticity ~\citep{10.1098/rsif.2018.0943}.

Existing methods \citep{vandegar2021neural,vetter2024sourcerer}  often build a differentiable surrogate of the simulator by sampling $\theta$ from a distribution $\pi(\theta)$ (we call this \textit{initial prior}) and running the simulator forward to generate training data. The prior is then optimized so that the resulting marginal matches $p_o(x)$ under the surrogate. However, when the surrogate is inaccurate in regions explored by the learned source, agreement under the surrogate need not imply agreement under the true simulator. 
Here, we instead propose an iterative approach \textbf{Posterior Averaging (PA)} where, at each round, we learn the posterior from data simulated under the current source estimate and construct the next source as the posterior averaged over the observed data. 
We show that this procedure implements an Expectation-Maximization scheme \citep{dempster1977maximum} whose induced-data KL decreases monotonically at every iteration in the idealized setting. 
If the iteration converges, the resulting source necessarily satisfies a \emph{self-consistency} condition with respect to the true likelihood: the source equals the average of the true posteriors it induces over real data. 
Since multiple source distributions can induce the same marginal, $q(\theta)$ is generally not identifiable from $p_o(x)$. We therefore evaluate methods by how well the recovered $q(\theta)$ reproduces the push-forwarded observation distribution, rather than by recovery of a single "true" source.

\section{Related Work}
\label{sec:related_works}

Under our non-differentiable black-box setting, the relevant implementations of existing SDE methods rely on a \textbf{fixed learned simulator surrogate} rather than direct simulator differentiation.

\paragraph{Neural Empirical Bayes (NEB)}
NEB \citep{vandegar2021neural} trains a normalizing flow $\hat p(x \mid \theta)$ as a likelihood surrogate. Training data is generated by fixing a broad prior $\pi(\theta)$, sampling $\theta \sim \pi(\theta)$, and simulating $x \sim p(x \mid \theta)$. $q_\phi(\theta)$ is then optimized to maximize $\sum_{x \sim p_o} \log q_\phi(x) = \sum_{x \sim p_o} \log \int \hat p(x \mid \theta)\, q_\phi(\theta)\, d\theta$.

\paragraph{Sample-based Maximum Entropy SDE (Sourcerer)}
Sourcerer  \citep{vetter2024sourcerer} resolves the non-uniqueness of SDE by selecting the maximum-entropy feasible source distribution:
$
    \max_\phi H(q_\phi)\,\, \text{s.t.} \,\, D(q_\phi, p_o) = 0,
$
where $D$ is a distance between distributions. Sourcerer's primary sample-based approach differentiates through the simulator directly; under our non-differentiable black-box assumption, only its surrogate-likelihood variant applies, so \emph{Sourcerer} in this paper refers to $q_\phi(x) = \int \hat p(x \mid \theta)\, q_\phi(\theta)\, d\theta$, with $\hat p(x\mid\theta)$ trained identically to NEB's surrogate.

\section{Method}
\label{sec:method}

Suppose some $\pi$ solves SDE exactly, $\int \pi(\theta)p(x\mid\theta)\,d\theta = p_o(x)$ for all $x$. Its true Bayes posterior is $\pi(\theta\mid x) = \pi(\theta)p(x\mid\theta)/p_o(x)$, and averaging over real data gives
$
\int p_o(x)\,\pi(\theta\mid x)\,dx = \int \pi(\theta)\,p(x\mid\theta)\,dx = \pi(\theta).
$
Therefore, while \cite{vetter2024sourcerer} claims that the average posterior does not always converge to a source distribution, \textbf{self-consistency} $\pi(\theta) = \int p_o(x)\,\pi(\theta\mid x)\,dx$ is indeed necessary for exact recovery. 
Since the objective of fixed-surrogate methods is stated purely in terms of a \emph{frozen} surrogate, any self-consistency they achieve is with respect to the surrogate, and only transfers to the true simulator if the surrogate is accurate over the fitted source. 

In our method, we alternate between training a posterior under the current source (E-step) and refitting the new source to the average posterior over real data (M-step), targeting self-consistency directly, and prove this alternation is monotone every round. 

\subsection{General algorithm}
\label{sec:general_alg}
Algorithm~\ref{alg:general} describes our method. At E-step, given an initial prior or the current source $\pi_t(\theta)$, we fit a posterior $p_t(\theta \mid x)$ via a Neural Posterior Estimator (NPE) \citep{papamakarios2016fast,greenberg2019automatic}
\[
\mathcal{L}_{\mathrm{NPE}}(\pi_t) = -\, \mathbb{E}_{\theta \sim \pi_t,\; x \sim p(x \mid \theta)}\big[\log p_t(\theta \mid x)\big].
\]
At M-step, given the posterior $p_t(\theta \mid x)$, the next source targets the average posterior over real data,
\[
\pi_{t+1}(\theta) = \int p_o(x)\, p_t(\theta \mid x)\, dx \;\approx\; \hat\pi_{t+1}(\theta) := \frac{1}{n} \sum_{i=1}^{n} p_t(\theta \mid x_i), \qquad x_i \in \mathcal{D},
\]
fit by $\mathcal{L}_{\mathrm{source}} = -\,\mathbb{E}_{\theta \sim \hat\pi_{t+1}}[\log \pi(\theta)]$ over pooled posterior samples. Next, we describe two variants of our method: \textbf{PA} with separate flows and \textbf{PA-Shared} with a shared flow.

\begin{algorithm}[ht]
\caption{General procedure}
\label{alg:general}
\begin{algorithmic}[1]
\Require Simulator $p(x\mid\theta)$, data $\mathcal{D}$, initial prior $\pi_0$, rounds $T$
\Repeat
    \State \textbf{E-step:} simulate $\theta_j \sim \pi_t$, $x_j \sim p(x\mid\theta_j)$; fit $p_t(\theta\mid x) \gets \arg\min \mathcal{L}_{\mathrm{NPE}}(\pi_t)$
    \State \textbf{M-step:} for each $x_i\in\mathcal D$, draw $\theta_s \sim p_t(\theta\mid x_i)$; fit $\pi_{t+1} \gets \arg\min \mathcal{L}_{\mathrm{source}}$
    \State $t \gets t+1$
\Until{$t=T$ \textbf{or} converged}
\end{algorithmic}
\end{algorithm}

\subsection{Variants}
\label{sec:variants}

\textbf{PA} (Algorithm~\ref{alg:pa}) represents the source via an unconditional flow $\pi_{\phi}(\theta)$, and the posterior via a conditional flow $p_{\psi}(\theta\mid x)$.
In practice, both source and posterior can be represented within a single conditional flow $f_\phi(z,c)$, where $c=e_\varnothing$ (a learnable null embedding) gives $\pi_\phi(\theta)=f_\phi(z,e_\varnothing)$, and $c=h_\xi(x)$ gives $p_\phi(\theta\mid x)=f_\phi(z,h_\xi(x))$. We refer to this implementation as \textbf{PA-Shared} (
(Algorithm~\ref{alg:pa-shared}). Context dropout during training ($c\to e_\varnothing$ with probability \ $p_\varnothing$, as in classifier-free guidance) lets one objective train both modes, so posterior training informs the prior's density estimate directly.
Since $\phi$ is shared, fitting the posterior mode also moves the source it represents. However, the context dropout during training can reduce this drift. This also requires drawing all M-step samples right after the E-step and before any M-step update.

\subsection{Convergence}
\label{sec:monotonicity}

Write $\pi_t(x) := \int p(x\mid\theta)\,\pi_t(\theta)\,d\theta$ for the marginal induced by $\pi_t$ through the simulator likelihood.

\begin{proposition}[Monotone $x$-marginal KL]
\label{prop:monotone}
Under (A1) full-support $\pi_t$ and (A2) exact E-/M-steps ($p_t(\theta\mid x)$ the exact posterior under $\pi_t$, and $\pi_{t+1}(\theta)=\int p_o(x)p_t(\theta\mid x)dx$ exactly), every round satisfies
\[
D_{\mathrm{KL}}\big(p_o(x) \,\|\,\pi_{t+1}(x)\big) \;\le\; D_{\mathrm{KL}}\big(p_o(x)\,\|\,\pi_t(x)\big),
\]
with equality iff $\pi_{t+1}=\pi_t$ a.e., provided the displayed quantities are finite. Full derivation in Appendix~\ref{app:kl-derivation}
\end{proposition}

\begin{remark}[Initialization]
\label{rem:init}
Monotonicity is informative only when $D_{\mathrm{KL}}(p_o\|\pi_0)<\infty$; in particular, if the initial induced marginal $\pi_0(x)$ assigns zero density to a region of positive $p_o$-mass, the initial KL is infinite and the bound is vacuous throughout.
\end{remark}

\subsection{Why Iteration Is Not Guaranteed to Help Fixed-Surrogate Baselines}
\label{sec:iteration-limits}

\paragraph{Our update targets self-consistency directly; the baselines target it only indirectly.}
Under (A2), our M-step \emph{is} the self-consistency operator $\pi_{t+1}(\theta) = \int p_o(x)\,p_t(\theta\mid x)\,dx$, with $p_t(\theta\mid x)$ the Bayes posterior induced by $\pi_t$ and the true likelihood $p(x\mid\theta)$; at a fixed point, $\pi_{t+1}=\pi_t$ reduces directly to the necessary condition above. In practice $p_t(\theta\mid x)$ is approximated by an NPE model $\hat p_t(\theta\mid x)$, giving
$
\hat\pi_{t+1}(\theta) = \frac{1}{n}\sum_{i=1}^n \hat p_t(\theta\mid x_i),
$
an approximation to, not an exact instance of, the self-consistency operator.

NEB and Sourcerer approximate a different object. Surrogate methods approximate $p(x\mid\theta)$ and then optimize an SDE objective under that approximation; our method instead approximates the Bayes posterior induced by the current source and true simulator samples, and therefore approximates the self-consistency operator directly. Self-consistency is thus not what the baselines' update evaluates but a byproduct, depending on the surrogate's accuracy and coverage. One could iterate this procedure: retrain the surrogate under the previously fitted source, then re-optimize the source, and repeat. But each round's objective is still stated purely in terms of that round's frozen surrogate rather than the self-consistency operator itself.
\section{Experiments}
\label{sec:experiments}
\subsection{Setup}
\label{sec:exp-setup}
\paragraph{Tasks and methods.}
We evaluate on three tasks
: \textbf{Two Moons} with $\theta\in\mathbb{R}^2$, \textbf{SLCP} with $\theta\in\mathbb{R}^5$, and \textbf{Lotka--Volterra}, \textbf{LV}, with $\theta\in\mathbb{R}^4$. We use their true priors as an exact source of each task to generate the observations. Exact task definitions are in Appendix~\ref{app:tasks}. Every method sees $n_{\text{obs}}=10{,}000$ observed datasets per task.

We compare PA and PA-Shared against baselines NEB and Sourcerer, and their iterated variants NEB-it and Sourcerer-it. There, in each round we re-simulate from the current source estimate, refit the surrogate from scratch, and refit the source, using the original authors' unmodified implementations. PA, PA-Shared, NEB-it, and Sourcerer-it run 8 rounds at 4,000 fresh simulator calls per round; NEB and Sourcerer fit one frozen surrogate on the same total budget of 32,000 simulations. We report data-space classifier two-sample test \citep{lopez-paz2017revisiting} (C2ST, closer to 0.5 is better, meaning two distributions are indistinguishable), mean $\pm$ std over 3 seeds. The architecture and optimization details of PA and PA-Shared are in Appendix~\ref{app:arch}.

\paragraph{Initial priors.}
We first study broad-support priors $\pi_0$ centred on or containing $\pi^\star$: \texttt{gaussian}, an unbounded diagonal Gaussian, and \texttt{widebox}, a uniform box with large margin (Section~\ref{sec:exp1}). Next, we examine misspecified priors with shifted support (Section~\ref{sec:exp2}). For this, we slide a box of $\pi^\star$'s own width right so it only partially overlaps $\pi^\star$, excluding a region of positive $\pi^\star$-mass: \texttt{shift\_half}, a quarter-width offset, and \texttt{shift\_full}, a half-width offset that leaves the box near one-sided. Table~\ref{tab:priors} gives exact specifications alongside $\pi^\star$.

\begin{table}[t]
\centering
\caption{$\pi_0$ specifications for all four variants, alongside the true prior $\pi^\star$.}
\label{tab:priors}
\small
\resizebox{0.9\textwidth}{!}{
\begin{tabular}{lccc}
\toprule
& Two Moons & SLCP & LV \\
\midrule
$\pi^\star$, true & $\mathrm{Unif}([-1,1]^2)$ & $\mathrm{Unif}([-3,3]^5)$ & $\alpha,\gamma\!\sim\!\mathrm{LogN}(-0.125,0.5)$, $\beta,\delta\!\sim\!\mathrm{LogN}(-3,0.5)$ \\
\midrule
\texttt{gaussian} & $\mathcal{N}(0,3^2I)$ & $\mathcal{N}(0,9^2I)$ & $\mathcal{N}((1.0,0.056,1.0,0.056),\mathrm{diag}(1.65,0.09,1.65,0.09)^2)$ \\
\texttt{widebox} & $[-10,10]^2$ & $[-10,10]^5$ & $[0,10]\times[0,1]\times[0,10]\times[0,1]$ \\
\texttt{shift\_half} & $[-0.5,1.5]^2$ & $[-1.5,4.5]^5$ & $[0.6,2.2]\times[0.033,0.123]\times[0.6,2.2]\times[0.033,0.123]$ \\
\texttt{shift\_full} & $[0,2]^2$ & $[0,6]^5$ & $[1.4,3.0]\times[0.08,0.17]\times[1.4,3.0]\times[0.08,0.17]$ \\
\bottomrule
\end{tabular}
}
\end{table}

\subsection{Broad initial prior}
\label{sec:exp1}

Table~\ref{tab:exp1} reports C2ST under the two broad-support variants. Our methods ranked best across nearly all task-prior combinations, with the exception of Two Moons, where NEB-it marginally outperformed our method: 0.513 vs.\ 0.515 under Gaussian, 0.498 vs.\ 0.524 under wide-box.
NEB-it improved substantially over one-shot NEB in every cell, for example on SLCP dropping from 0.894 to 0.646 under \texttt{gaussian} and from 0.650 to 0.611 under \texttt{widebox}. Meanwhile Sourcerer-it was worse than one-shot Sourcerer on SLCP under both priors, rising from 0.844 to 0.934 and from 0.855 to 0.922, respectively.
LV was the hardest task where most methods failed. PA-Shared clearly performed best at 0.655 and 0.675 for \texttt{gaussian} and \texttt{widebox}, respectively. PA performed worse (0.914 and 0.892), but still outperformed all baselines on both priors ($>0.964$).

\begin{table}[t]
\centering
\caption{Data-space C2ST under broad prior \texttt{gaussian} and \texttt{widebox} $\pi_0$.}
\label{tab:exp1}
\resizebox{\textwidth}{!}{
\begin{tabular}{lcccccc}
\toprule
& \multicolumn{3}{c}{\texttt{gaussian}}
& \multicolumn{3}{c}{\texttt{widebox}} \\
\cmidrule(lr){2-4} \cmidrule(lr){5-7}
Method
& Two Moons & SLCP & LV
& Two Moons & SLCP & LV \\
\midrule
NEB
& $0.514 \pm 0.013$
& $0.894 \pm 0.037$
& $0.993 \pm 0.004$
& $0.588 \pm 0.034$
& $0.650 \pm 0.041$
& $0.964 \pm 0.007$ \\

Sourcerer
& $0.588 \pm 0.013$
& $0.844 \pm 0.035$
& $1.000 \pm 0.000$
& $0.601 \pm 0.013$
& $0.855 \pm 0.022$
& $0.999 \pm 0.001$ \\
\midrule
NEB-it
& $\mathbf{0.513 \pm 0.002}$
& $0.646 \pm 0.006$
& $0.987 \pm 0.008$
& $\mathbf{0.498 \pm 0.010}$
& $0.611 \pm 0.016$
& $0.992 \pm 0.004$ \\

Sourcerer-it
& $0.572 \pm 0.007$
& $0.934 \pm 0.013$
& $1.000 \pm 0.000$
& $0.591 \pm 0.009$
& $0.922 \pm 0.010$
& $1.000 \pm 0.000$ \\
\midrule
PA
& $0.515 \pm 0.008$
& $\mathbf{0.577 \pm 0.011}$
& $0.914 \pm 0.008$
& $0.524 \pm 0.014$
& $0.547 \pm 0.015$
& $0.892 \pm 0.017$ \\

PA-Shared
& $0.521 \pm 0.024$
& $0.584 \pm 0.022$
& $\mathbf{0.655 \pm 0.027}$
& $0.551 \pm 0.043$
& $\mathbf{0.530 \pm 0.007}$
& $\mathbf{0.675 \pm 0.034}$ \\
\bottomrule
\end{tabular}
}
\end{table}

\subsection{Misspecified initial prior}
\label{sec:exp2}

Table~\ref{tab:exp2} reports C2ST under the two misspecified prior settings. On Two Moons, NEB-it again performed marginally best, 0.497 vs.\ 0.521 for PA under \texttt{shift\_half} and 0.502 vs.\ 0.516 under \texttt{shift\_full}. 
On the other two tasks, iteration was unreliable for both baselines, more so than under the broad priors: NEB-it now lost to one-shot NEB on SLCP in both conditions, rising from 0.545 to 0.586 under \texttt{shift\_half} and from 0.592 to 0.612 under \texttt{shift\_full}. Sourcerer-it again performed substantially worse than one-shot Sourcerer on SLCP, rising from 0.584 to 0.913 under \texttt{shift\_half} and from 0.712 to 0.915 under \texttt{shift\_full}.
In LV, under \texttt{shift\_half}, both our methods clearly outperformed the baselines: PA at 0.683 and PA-Shared at 0.643. Under \texttt{shift\_full}, PA reached 0.903 while PA-Shared failed at 0.984, similarly to the baselines.

\begin{table}[t]
\centering
\caption{Data-space C2ST under mild \texttt{shift\_half} and near one-sided \texttt{shift\_full} misspecification.}
\label{tab:exp2}
\resizebox{\textwidth}{!}{%
\begin{tabular}{lcccccc}
\toprule
& \multicolumn{3}{c}{\texttt{shift\_half}}
& \multicolumn{3}{c}{\texttt{shift\_full}} \\
\cmidrule(lr){2-4} \cmidrule(lr){5-7}
Method
& Two Moons & SLCP & LV
& Two Moons & SLCP & LV \\
\midrule
NEB
& $0.513 \pm 0.007$
& $\mathbf{0.545 \pm 0.026}$
& $0.979 \pm 0.008$
& $0.526 \pm 0.008$
& $0.592 \pm 0.014$
& $0.987 \pm 0.001$ \\

Sourcerer
& $0.525 \pm 0.013$
& $0.584 \pm 0.021$
& $0.990 \pm 0.002$
& $0.536 \pm 0.013$
& $0.712 \pm 0.032$
& $0.993 \pm 0.001$ \\
\midrule
NEB-it
& $\mathbf{0.497 \pm 0.007}$
& $0.586 \pm 0.008$
& $0.986 \pm 0.009$
& $\mathbf{0.502 \pm 0.006}$
& $0.612 \pm 0.009$
& $0.992 \pm 0.007$ \\

Sourcerer-it
& $0.528 \pm 0.011$
& $0.913 \pm 0.007$
& $0.995 \pm 0.004$
& $0.528 \pm 0.011$
& $0.915 \pm 0.014$
& $0.998 \pm 0.001$ \\
\midrule
PA
& $0.521 \pm 0.012$
& $0.589 \pm 0.007$
& $0.683 \pm 0.017$
& $0.516 \pm 0.009$
& $0.713 \pm 0.008$
& $\mathbf{0.903 \pm 0.038}$ \\

PA-Shared
& $0.529 \pm 0.004$
& $0.561 \pm 0.005$
& $\mathbf{0.643 \pm 0.010}$
& $0.524 \pm 0.015$
& $\mathbf{0.586 \pm 0.021}$
& $0.984 \pm 0.005$ \\
\bottomrule
\end{tabular}%
}
\end{table}

\subsection{Convergence across rounds}
\label{sec:convergence}

\begin{figure}
    \centering
    \includegraphics[width=\linewidth]{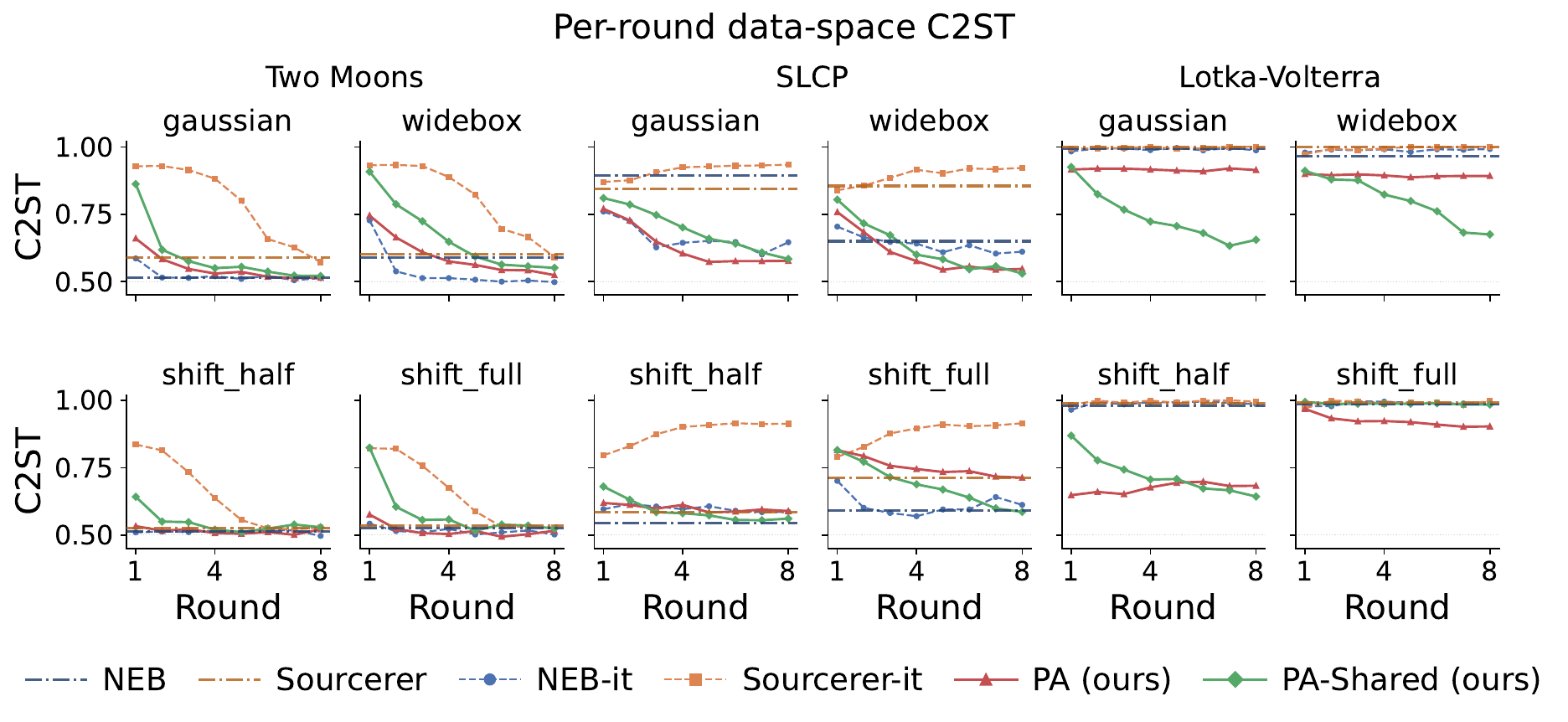}
    \caption{Data-space C2ST by round for the four iterative methods: NEB-it, Sourcerer-it, PA, and PA-Shared. One-shot baselines (flat by construction) are included for reference.}
    \label{fig:C2ST_iterative}
\end{figure}

Figure~\ref{fig:C2ST_iterative} plots per-round C2ST for the four iterative methods. NEB-it and both our methods generally descended across rounds. In contrast, Sourcerer-it improved for Two Moons, stayed constant for LV, but deteriorated over rounds for the SLCP task. 

\section{Discussion and Conclusion}
\label{sec:discussion}

We started from the observation that existing SDE methods state their objective purely in terms of a surrogate likelihood trained once under a fixed proposal prior, so agreement under the surrogate need not imply agreement under the true simulator. Posterior averaging sidesteps this by never forming a surrogate likelihood at all: each round trains a posterior on fresh draws from the current source and refits the source to that posterior averaged over the observed data, which we prove decreases the induced-data KL monotonically under idealized conditions. Empirically, it consistently outperforms fixed-surrogate baselines and their iterated variants, most clearly on Lotka--Volterra, where iteration alone does not close the surrogate gap.

\paragraph{Limitations.} Because multiple sources induce the same marginal, our evaluation speaks to reproducing $p_o$ and not to recovering $\pi^\star$: a source scoring well in data space may still place mass where a domain scientist would not. \cref{prop:monotone} assumes exact E- and M-steps, whereas both are neural approximations in practice, and \cref{rem:init} makes it vacuous whenever $D_{\mathrm{KL}}(p_o\|\pi_0)=\infty$. Our black-box assumption excludes Sourcerer's primary sample-based variant,  which differentiates through the simulator directly; so we compare only against its surrogate-likelihood form. Finally, we evaluate on three benchmark simulators with synthetic $p_o$, a single simulation budget, $n=10{,}000$ observations, and three seeds; behaviour on real observations, at smaller $n$, and on higher-dimensional simulators remains open.

\paragraph{Concurrent work: Simulation-Based Empirical Bayes (SBEB)}

Concurrently with this work, \citet{shen2026simulationbasedempiricalbayes} propose a closely related method with an equivalent convergence guarantee, matching our own monotonicity result (Proposition~\ref{prop:monotone}). While we make no priority claim relative to this work, we introduce some differences: SBEB is framed as simulation-based empirical Bayes and represents the fitted population prior implicitly through posterior samples, whereas our SDE formulation additionally fits an explicit queryable source density and studies its relation to existing SDE baselines, both fixed-surrogate and iterative (Section~\ref{sec:iteration-limits}).

\section*{Acknowledgments.} 

This project was supported by the Diabetes Center Berne and strategic funding of the medical faculty of the University of Bern. Calculations were performed on UBELIX, the HPC cluster at the University of Bern.
\clearpage

\bibliographystyle{plainnat}
\bibliography{neurips_2026}

@misc{shen2026simulationbasedempiricalbayes, 
title={Simulation-Based Empirical Bayes}, 
author={Xinwei Shen and Diana Cai and Cheng Zhang and David M. Blei}, 
year={2026}, 
eprint={2607.21843}, 
archivePrefix={arXiv}, 
primaryClass={stat.ML}, 
url={https://arxiv.org/abs/2607.21843}, }

@inproceedings{
vetter2024sourcerer,
title={Sourcerer: Sample-based Maximum Entropy Source Distribution Estimation},
author={Julius Vetter and Guy Moss and Cornelius Schr{\"o}der and Richard Gao and Jakob H. Macke},
booktitle={The Thirty-eighth Annual Conference on Neural Information Processing Systems},
year={2024},
url={https://openreview.net/forum?id=0cgDDa4OFr}
}

@inproceedings{vandegar2021neural,
	AUTHOR = {Vandegar, Maxime and Kagan, Michael and Wehenkel, Antoine and Louppe, Gilles},
	EPRINT = {https://orbi.uliege.be/handle/2268/253953},
	EPRINTTYPE = {hdl},
	TITLE = {Neural Empirical Bayes: Source Distribution Estimation and its Applications to Simulation-Based Inference},
	LANGUAGE = {English},
	YEAR = {2021},
	BOOKTITLE = {Proceedings of AISTATS 2021},
	URL = {https://arxiv.org/abs/2011.05836}
}

@article{YAO2026140531,
title = {Efficient probabilistic inference for expensive-to-evaluate black-box models for full-scale nuclear reactor thermal hydraulics},
journal = {Energy},
volume = {348},
pages = {140531},
year = {2026},
issn = {0360-5442},
doi = {https://doi.org/10.1016/j.energy.2026.140531},
url = {https://www.sciencedirect.com/science/article/pii/S0360544226006341},
author = {Zikang Yao and Eugene Yee and Fue-Sang Lien},
}

@article{10.1098/rsif.2018.0943,
    author = {Warne, David J. and Baker, Ruth E. and Simpson, Matthew J.},
    title = {Simulation and inference algorithms for stochastic biochemical reaction networks: from basic concepts to state-of-the-art},
    journal = {Journal of The Royal Society Interface},
    volume = {16},
    number = {151},
    pages = {20180943},
    year = {2019},
    month = {02},
    issn = {1742-5689},
    doi = {10.1098/rsif.2018.0943},
    url = {https://doi.org/10.1098/rsif.2018.0943},
    eprint = {https://royalsocietypublishing.org/rsif/article-pdf/doi/10.1098/rsif.2018.0943/890028/rsif.2018.0943.pdf},
}

@inproceedings{
lopez-paz2017revisiting,
title={Revisiting Classifier Two-Sample Tests},
author={David Lopez-Paz and Maxime Oquab},
booktitle={International Conference on Learning Representations},
year={2017},
url={https://openreview.net/forum?id=SJkXfE5xx}
}

@InProceedings{lueckmann2021benchmarking,
 title     = {Benchmarking Simulation-Based Inference},
 author    = {Lueckmann, Jan-Matthis and Boelts, Jan and Greenberg, David and Goncalves, Pedro and Macke, Jakob},
 booktitle = {Proceedings of The 24th International Conference on Artificial Intelligence and Statistics},
 pages     = {343--351},
 year      = {2021},
 editor    = {Banerjee, Arindam and Fukumizu, Kenji},
 volume    = {130},
 series    = {Proceedings of Machine Learning Research},
 month     = {13--15 Apr},
 publisher = {PMLR}
}

@article{Stimper2023, 
  author = {Vincent Stimper and David Liu and Andrew Campbell and Vincent Berenz and Lukas Ryll and Bernhard Schölkopf and José Miguel Hernández-Lobato}, 
  title = {normflows: A PyTorch Package for Normalizing Flows}, 
  journal = {Journal of Open Source Software}, 
  volume = {8},
  number = {86}, 
  pages = {5361}, 
  publisher = {The Open Journal}, 
  doi = {10.21105/joss.05361}, 
  url = {https://doi.org/10.21105/joss.05361}, 
  year = {2023}
}

@article{BoeltsDeistler_sbi_2025,
  doi = {10.21105/joss.07754},
  url = {https://doi.org/10.21105/joss.07754},
  year = {2025},
  publisher = {The Open Journal},
  volume = {10},
  number = {108},
  pages = {7754},
  author = {Jan Boelts and Michael Deistler and Manuel Gloeckler and Álvaro Tejero-Cantero and Jan-Matthis Lueckmann and Guy Moss and Peter Steinbach and Thomas Moreau and Fabio Muratore and Julia Linhart and Conor Durkan and Julius Vetter and Benjamin Kurt Miller and Maternus Herold and Abolfazl Ziaeemehr and Matthijs Pals and Theo Gruner and Sebastian Bischoff and Nastya Krouglova and Richard Gao and Janne K. Lappalainen and Bálint Mucsányi and Felix Pei and Auguste Schulz and Zinovia Stefanidi and Pedro Rodrigues and Cornelius Schröder and Faried Abu Zaid and Jonas Beck and Jaivardhan Kapoor and David S. Greenberg and Pedro J. Gonçalves and Jakob H. Macke},
  title = {sbi reloaded: a toolkit for simulation-based inference workflows},
  journal = {Journal of Open Source Software}
}

@article{sbi_frontier,
author = {Kyle Cranmer  and Johann Brehmer  and Gilles Louppe },
title = {The frontier of simulation-based inference},
journal = {Proceedings of the National Academy of Sciences},
volume = {117},
number = {48},
pages = {30055-30062},
year = {2020},
doi = {10.1073/pnas.1912789117},
URL = {https://www.pnas.org/doi/abs/10.1073/pnas.1912789117},
eprint = {https://www.pnas.org/doi/pdf/10.1073/pnas.1912789117}}

@article{dempster1977maximum,
 ISSN = {00359246},
 URL = {http://www.jstor.org/stable/2984875},
 author = {A. P. Dempster and N. M. Laird and D. B. Rubin},
 journal = {Journal of the Royal Statistical Society. Series B (Methodological)},
 number = {1},
 pages = {1--38},
 publisher = {[Royal Statistical Society, Oxford University Press]},
 title = {Maximum Likelihood from Incomplete Data via the EM Algorithm},
 urldate = {2026-09-01},
 volume = {39},
 year = {1977}
}

@inproceedings{papamakarios2016fast,
 author = {Papamakarios, George and Murray, Iain},
 booktitle = {Advances in Neural Information Processing Systems},
 editor = {D. Lee and M. Sugiyama and U. Luxburg and I. Guyon and R. Garnett},
 pages = {},
 publisher = {Curran Associates, Inc.},
 title = {Fast $\epsilon$-free Inference of Simulation Models with Bayesian Conditional Density Estimation},
 url = {https://proceedings.neurips.cc/paper_files/paper/2016/file/6aca97005c68f1206823815f66102863-Paper.pdf},
 volume = {29},
 year = {2016}
}

@inproceedings{greenberg2019automatic,

        editor = {Chaudhuri, Kamalika;Salakhutdinov, Ruslan},
	author = {Greenberg, David and Nonnenmacher, Marcel and Macke, Jakob},
	title = {Automatic Posterior Transformation for Likelihood-Free Inference},
	booktitle = {Proceedings of the 36th International Conference on Machine Learning},
	year = {2019},

        month = {09--15 Jun},

        volume = {97},

        publisher = {PMLR},
        series = {Proceedings of Machine Learning Research},
        url = {http://proceedings.mlr.press/v97/greenberg19a.html},
}

\clearpage

\appendix

\section{Convergence Proof}
\label{app:kl-derivation}

\subsection{Full KL Derivation}

\begin{align}
D_{\mathrm{KL}}\big(p_o(x) \,\|\, \pi_{t+1}(x)\big)
&= -H[p_o(x)] - \int p_o(x) \log \pi_{t+1}(x)\, dx \label{eq:step1}\\
&= -H[p_o(x)] - \int p_o(x) \log\!\left( \int p_t(\theta \mid x)\, \frac{\pi_{t+1}(\theta)\, p(x \mid \theta)}{p_t(\theta \mid x)}\, d\theta \right) dx \label{eq:step3}\\
&\leq -H[p_o(x)] - \iint p_o(x)\, p_t(\theta \mid x)\, \log\!\left( \frac{\pi_{t+1}(\theta)\, p(x \mid \theta)}{p_t(\theta \mid x)} \right) d\theta \, dx \label{eq:jensen}\\
&= -H[p_o(x)] - \iint p_o(x)\,p_t(\theta\mid x)\Big[\log \pi_{t+1}(\theta) + \log p(x\mid\theta) - \log p_t(\theta \mid x)\Big] d\theta\,dx \label{eq:split}\\
&= -H[p_o(x)] - \int \pi_{t+1}(\theta) \log \pi_{t+1}(\theta)\, d\theta \notag\\
&\qquad - \iint p_o(x)\,p_t(\theta\mid x)\Big[\log p(x\mid\theta) - \log p_t(\theta \mid x)\Big] d\theta\,dx \label{eq:fubini1}\\
&\leq -H[p_o(x)] - \int \pi_{t+1}(\theta) \log \pi_{t}(\theta)\, d\theta \notag\\
&\qquad - \iint p_o(x)\,p_t(\theta\mid x)\Big[\log p(x\mid\theta) - \log p_t(\theta \mid x)\Big] d\theta\,dx \label{eq:gibbs}\\
&= -H[p_o(x)] - \iint p_o(x)\,p_t(\theta\mid x)\Big[\log \pi_t(\theta) + \log p(x\mid\theta) - \log p_t(\theta \mid x)\Big] d\theta\,dx \label{eq:fubini2}\\
&= -H[p_o(x)] - \iint p_o(x)\,p_t(\theta\mid x)\, \log \pi_t(x)\, d\theta\,dx \label{eq:posterior-identity}\\
&= -H[p_o(x)] - \int p_o(x) \log \pi_t(x)\, dx \;=\; D_{\mathrm{KL}}\big(p_o(x) \,\|\, \pi_t(x)\big). \label{eq:step8}
\end{align}

\subsection{Well-Definedness and Equality Condition}

\emph{Well-definedness of \eqref{eq:step3}:} wherever $\pi_{t+1}(\theta)p(x\mid\theta)>0$, both factors are positive, so $p(x\mid\theta)>0$; by (A1), $\pi_t(\theta)>0$ and $\pi_t(x)>0$, so $p_t(\theta\mid x)>0$ exactly where needed; no support relation between $\pi_{t+1}$ and $\pi_t$ beyond (A1) is required.

\eqref{eq:jensen}: Jensen on concave $\log$ under $p_t(\theta\mid x)\,d\theta$. \eqref{eq:split}: splits the log-product before integrating. \eqref{eq:fubini1} and \eqref{eq:fubini2}: Fubini, via (A2), to move between $\int p_o(x)p_t(\theta\mid x)\,dx$ and $\pi_{t+1}(\theta)$. \eqref{eq:gibbs} applies Gibbs' inequality, using that $\pi_{t+1}$ maximizes $\int\pi_{t+1}\log q$ over $q$, so swapping in $\log\pi_t$ cannot decrease the bound. \eqref{eq:posterior-identity}: the exact-posterior identity $\pi_t(\theta)p(x\mid\theta)/p_t(\theta\mid x)=\pi_t(x)$, constant in $\theta$.

\paragraph{Equality condition.} $D_{\mathrm{KL}}(p_o\,\|\,\pi_{t+1}) = D_{\mathrm{KL}}(p_o\,\|\,\pi_t)$ iff $\pi_{t+1} = \pi_t$ a.e., by strictness of Gibbs' inequality unless the two priors agree, provided the displayed KL quantities are finite.

\section{Algorithm Details}
\label{app:algorithms}

\subsection{PA (Separate Flows)}
\label{app:alg-pa}

\begin{algorithm}[H]
\caption{PA with separate flows}
\label{alg:pa}
\begin{algorithmic}[1]
\Require Simulator $p(x\mid\theta)$, data $\mathcal{D}$, initial prior $\phi^{(0)}$, rounds $T$
\Repeat
    \State \textbf{E-step:} simulate $\theta_j \sim \pi_{\phi^{(t)}}$, $x_j \sim p(x\mid\theta_j)$; $\psi^{(t+1)} \gets \arg\min_\psi \mathcal{L}_{\mathrm{NPE}}(\psi;\pi_{\phi^{(t)}})$
    \State \textbf{M-step:} for each $x_i\in\mathcal D$, draw $\theta_s \sim p_{\psi^{(t+1)}}(\theta\mid x_i)$; $\phi^{(t+1)} \gets \arg\min_\phi \mathcal{L}_{\mathrm{source}}(\phi)$
    \State $t \gets t+1$
\Until{$t=T$ \textbf{or} converged}
\end{algorithmic}
\end{algorithm}

\subsection{PA-Shared (Shared Flow)}
\label{app:alg-PA-Shared}

\begin{algorithm}[H]
\caption{PA with shared flow}
\label{alg:pa-shared}
\begin{algorithmic}[1]
\Require Simulator $p(x\mid\theta)$, data $\mathcal D$, initial $\phi^{(0)}$, dropout $p_\varnothing$, rounds $T$
\Repeat
    \State \textbf{E-step:} simulate $\theta_j\sim\pi_{\phi^{(t)}}$, $x_j\sim p(x\mid\theta_j)$, apply CFG dropout; $(\phi_{\mathrm E}^{(t+1)},\xi^{(t+1)}) \gets \arg\min \mathcal L_{\mathrm E}(\phi,\xi)$
    \State \textbf{M-step:} embed $e_i = h_{\xi^{(t+1)}}(x_i)$; draw $\theta_s \sim f_{\phi_{\mathrm E}^{(t+1)}}(z,e_i)$ \Comment{frozen E-step output, before M-step's own gradient}
    \State $\phi^{(t+1)} \gets \arg\min_\phi \mathcal L_{\mathrm{prior}}(\phi)$, initialized from $\phi_{\mathrm E}^{(t+1)}$, at $c=e_\varnothing$; \quad $t\gets t+1$
\Until{$t=T$ \textbf{or} converged}
\end{algorithmic}
\end{algorithm}

\section{Task Definitions}
\label{app:tasks}

We utilize three tasks from the \texttt{sbibm} simulation-based inference benchmark suite \citep{lueckmann2021benchmarking}. We reuse the corresponding simulators and treat their true priors as one of the true source distributions.

\subsection{Two Moons}

A two-dimensional task whose true prior is $\mathrm{Unif}(-1,1)$ per dimension. Given $\theta = (\theta_1,\theta_2)$, data are generated by combining a fixed-radius arc with a translation depending on $\theta$:
\[
x \mid \theta =
\begin{pmatrix}
r\cos\alpha + 0.25 \\
r\sin\alpha
\end{pmatrix}
+
\begin{pmatrix}
-\lvert \theta_1+\theta_2\rvert/\sqrt{2} \\
(-\theta_1+\theta_2)/\sqrt{2}
\end{pmatrix},
\qquad \alpha \sim \mathrm{Unif}(-\tfrac{\pi}{2},\tfrac{\pi}{2}),\;\; r\sim\mathcal{N}(0.1,0.01^2).
\]
$\theta \in \mathbb{R}^2$, $x \in \mathbb{R}^2$.

\subsection{SLCP}

A five-dimensional task whose true prior is $\mathrm{Unif}(-3,3)$ per dimension. The observation is four i.i.d.\ draws from a 2-D Gaussian whose mean and covariance are nonlinear functions of $\theta$:
\[
x = (x_1,\dots,x_4),\quad x_i \sim \mathcal{N}(m_\theta, S_\theta), \qquad
m_\theta = \begin{pmatrix}\theta_1\\\theta_2\end{pmatrix}, \quad
S_\theta = \begin{pmatrix} s_1^2 & \rho s_1 s_2 \\ \rho s_1 s_2 & s_2^2 \end{pmatrix},
\]
with $s_1=\theta_3^2$, $s_2=\theta_4^2$, $\rho=\tanh\theta_5$. Concatenating the four draws gives $x \in \mathbb{R}^8$.

\subsection{Lotka--Volterra}

A predator-prey ODE model, included as the one real (non-toy) simulator in our evaluation. Four rate parameters $\theta = (\alpha,\beta,\gamma,\delta)$ govern
\[
\frac{dX}{dt} = \alpha X - \beta XY, \qquad \frac{dY}{dt} = -\gamma Y + \delta XY,
\]
with fixed initial condition $(X(0),Y(0))=(30,1)$ and integration horizon $T=20$. The true (\texttt{sbibm}) prior is
\[
\alpha, \gamma \sim \mathrm{LogNormal}(-0.125, 0.5), \quad \beta, \delta  \sim \mathrm{LogNormal}(-3, 0.5)
\]
Observations are noisy log-population counts at 10 evenly spaced time points for each species, $x_{1,i}\sim\mathrm{LogNormal}(\log X_i, 0.1)$, $x_{2,i}\sim\mathrm{LogNormal}(\log Y_i,0.1)$, giving $x \in \mathbb{R}^{20}$. 

\section{Architecture and Optimization}
\label{app:arch}

All hyperparameters are fixed across tasks, seeds, and both experiments unless noted.

\subsection{PA (Separate Flows)}

\begin{itemize}
  \item \textbf{Disjoint parameters} for the source $p_\phi(\theta)$ and the posterior $q_\psi(\theta\mid x)$.
  \item \textbf{Source flow $p_\phi$:} \texttt{normflows} \citep{Stimper2023}, diagonal-Gaussian base, $6\times$ [autoregressive rational-quadratic spline coupling (2 bins, hidden 64) + LU linear permute].
  \item \textbf{Posterior $q_\psi$:} \texttt{sbi} \citep{BoeltsDeistler_sbi_2025} NPE with the \texttt{nsf} density estimator. Training pairs are drawn from $p_\phi$.
  \item \textbf{E-step:} sample $n_{\text{sim}}$ parameters $\theta\sim p_\phi$, simulate, retrain the NPE (batch size of 256, max epochs of 50, early-stop patience 10).
  \item \textbf{M-step:} target $\hat p(\theta)=\tfrac1N\sum_i q_\psi(\theta\mid x_i^{\text{real}})$; draw 300 posterior samples per $x_i^{\text{real}}$ and fit $p_\phi$ by maximum likelihood (300 steps, batch 512, gradient-norm clip 5).
\end{itemize}

\subsection{PA-Shared (Shared Flow)}

\begin{itemize}
  \item A \textbf{single} conditional flow $\phi$: the same base and $6$-block
        structure as the source flow of \textbf{PA}, but each spline coupling additionally consumes a 32-dim
        context.
  \item $p_\phi(\theta)=\mathrm{flow}(z,\,c{=}e_\varnothing)$ and
        $q_\phi(\theta\mid x)=\mathrm{flow}(z,\,c{=}h_\xi(x))$, with
        $e_\varnothing\in\mathbb{R}^{32}$ a learnable null embedding
        (init.\ $\mathcal{N}(0,0.02^2)$).
  \item \textbf{Embedder $h_\xi$:} MLP $d_x\!\to\!64\!\to\!64\!\to\!32$, SiLU
        (a CNN for image-valued $x$). Inputs are $z$-scored per feature using
        the statistics of $\{x_i^{\text{real}}\}$.
  \item \textbf{Classifier-free guidance:} during the E-step the context is
        replaced by $e_\varnothing$ with probability $p_{\text{uncond}}=0.2$; at
        inference $c=e_\varnothing+w\,(h_\xi(x)-e_\varnothing)$ with $w=1$
        throughout.
  \item \textbf{E-step:} sample from the null-context branch, simulate, train $\phi$ and $\xi$ jointly on the conditional NLL of $\theta\mid h_\xi(x)$ (batch size of 256, max epochs of 60, patience 10, gradient-norm clip 5).
  \item \textbf{M-step:} target $p_\phi(\theta)=\mathbb{E}_{x^{\text{real}}}[q_\phi(\theta\mid x)]$; fit the null-context branch by maximum likelihood (300 steps, batch size of 256) on 100 posterior samples per $x_i^{\text{real}}$.
  \item Shared $\phi$ $\Rightarrow$ snapshot $\phi^{(t)}$ before the E-step, draw
        the M-step samples from the frozen $\phi^{(t)}$ combined with the updated
        embedder $\xi^{(t+1)}$, then restore the E-step-updated weights before
        the M-step gradient steps.
\end{itemize}

\subsection{Optimization Configuration}

\begin{table}[H]
\centering
\caption{Optimization configuration for PA and PA-Shared.}
\label{tab:opt-config}
\small
\begin{tabular}{@{}l p{4.4cm} p{4.4cm}@{}}
\toprule
 & PA & PA-Shared \\
\midrule
Init pre-fit to round-0 prior      & 800 steps, Adam 3e-3, batch 256 & 800 steps, Adam 3e-3, batch 256 \\
E-step optimizer                   & \texttt{sbi} NPE (Adam), batch 256, $\le$50 ep, patience 10 & Adam 1e-3 (flow+embedder), batch 256, $\le$60 ep, patience 10 \\
M-step optimizer                   & Adam 1e-3, 300 steps, batch 512 & Adam 1e-3 (shared), 300 steps, batch 256 \\
\bottomrule
\end{tabular}
\end{table}

\subsection{Protocol}

\begin{itemize}
  \item \textbf{Budget (matched, all methods):} $T=8$ rounds; 4000 $(\theta,x)$ pairs per E-step; 32{,}000 total. No accumulation across rounds --- each E-step uses fresh draws from the current source.
  \item \textbf{Observed set:} $N=10{,}000$, drawn once from the true prior.
  \item \textbf{Tasks (\texttt{sbibm}):} \texttt{two\_moons} ($d_\theta{=}2,\ d_x{=}2$), \texttt{slcp} ($5,\ 8$), \texttt{lotka\_volterra} ($4,\ 20$). \textbf{Seeds:} 0, 1, 2.
  \item \textbf{Metric:} data-space C2ST per round (classifier two-sample test, 1000 samples per side, between $x$ simulated from the round-$t$ learned source and $x^{\text{real}}$; 0.5 = indistinguishable). We report the performance in the last round and aggregate it over the 3 seeds.
  \item \textbf{Compute:} a single RTX 4090; the simulator runs on CPU.
\end{itemize}

\newpage
\end{document}